\documentclass[final,1p,times,authoryear]{elsarticle}

\journal{Artificial Intelligence}

\usepackage{amssymb}
\usepackage{amsthm}

\usepackage{booktabs}
\usepackage{tabularx}
\usepackage{enumitem}
\usepackage[x11names]{xcolor}
\usepackage[colorlinks]{hyperref}

\AtBeginDocument{\hypersetup{allcolors=DodgerBlue4}}

\newcommand{\vbarmc}[2]{\multicolumn{#1}{|c}{#2}}

\makeatletter
\newcommand*{\centerfloat}{%
  \parindent \z@
  \leftskip \z@ \@plus 1fil \@minus \textwidth
  \rightskip\leftskip
  \parfillskip \z@skip}
\makeatother

\makeatletter
\def\ps@pprintTitle{%
 \let\@oddhead\@empty
 \let\@evenhead\@empty
 \def\@oddfoot{}%
 \let\@evenfoot\@oddfoot}
\makeatother

\newcommand{\possessivecite}[1]{\citeauthor{#1}'s (\citeyear{#1})}

\usepackage{floatpag}

\usepackage{caption}
\setlist[description]{font=\normalfont\itshape\textbullet\space}

\theoremstyle{remark}
\newtheorem{mytheorem}{Theorem}

\usepackage{tikz-network}

\usepackage{longtable} 
\usepackage{pdflscape} 

\usepackage{placeins}

\usepackage{afterpage}

\usepackage[T1]{fontenc}

\begin{document}

\begin{frontmatter}

\title{Manipulation and Peer Mechanisms: A Survey}

\author{Matthew Olckers}

\address{Macquarie University and the e61 Institute}

\author{Toby Walsh}

\address{School of Computer Science and Engineering, UNSW Sydney}


\begin{abstract}
In peer mechanisms, the competitors for a prize also determine who wins. Each competitor may be asked to rank, grade, or nominate peers for the prize. Since the prize can be valuable, such as financial aid, course grades, or an award at a conference, competitors may be tempted to manipulate the mechanism. We survey approaches to prevent or discourage the manipulation of peer mechanisms. We conclude our survey by identifying several important research challenges.
\end{abstract}


\end{frontmatter}

\section{Introduction}

Imagine that you are competing for a prize at your workplace. The prize is awarded by asking everyone at your work, including you, to nominate who deserves the prize. The person with the most nominations wins. Who should you nominate? If you tell the truth and nominate who you think is most deserving, your nomination could cause you to lose out. Would you be truthful? Do you think your colleagues would be truthful?

As this example illustrates, when the competitors for a prize also determine who wins, the competitors may be tempted to manipulate the outcome. A growing research literature, which we collect under the term ``peer mechanisms'', aims to prevent or discourage manipulation in these situations. This paper surveys both the theoretical and empirical research on peer mechanisms and offers several research challenges. 

The interest in peer mechanisms has been fueled by the variety of high-stakes applications. The prize can take many forms. Closest to the experience of researchers, the prize could be an award at a conference. Further afield, the prize could be grades in a course \citep{topping1998peergrading}, a time slot to use a telescope \citep{merrifield2009telescope}, aid targeted to people in need \citep{conning2002community, alatas2012targeting}, loans for entrepreneurs \citep{hussam2021targeting}, a job for freelancers \citep{kotturi2020hirepeer}, an award for the best soccer player of the year \citep{caragiannis2019additive}, or even the papacy of the Catholic Church \citep{mackenzie2020pope}.

The variety of applications has inspired a variety of models. Some models, known as \emph{peer selection}, assume the mechanism designer wishes to select a single participant or a limited number of participants for the prize. While other models, known as \emph{peer grading}, assume each participant should receive a cardinal score. Besides the mechanism's output, the models can differ in many other dimensions, such as what the participants are asked to report, the type of information participants hold about their peers, and whether the mechanism designer can make payments. 

We provide a taxonomy to highlight the differences in each model introduced in the literature, and to identify some common themes. The approaches to prevent manipulation in peer mechanisms can be grouped into one of three categories:
\begin{enumerate}
    \item \emph{impartial} mechanisms where a participant's report cannot impact their chance of winning the prize,
    \item \emph{audits} where the mechanism attempts to detect and punish manipulation,
    \item \emph{rewards} for truthful reports.
\end{enumerate}
Although these three approaches are distinct, they are not mutually exclusive. The approaches
can be combined. The context will determine which approach or combination of approaches is
most suitable.

The bulk of the theoretical research has focused on impartial mechanisms. We delve deeper into these results by focusing on the model of peer selection, where peers nominate each other for a single prize or several identical prizes. Researchers have taken two main approaches: checking if impartiality is compatible with desirable axioms and checking how closely impartial mechanisms can approximate the most desirable outcome (such as awarding the prize to the participants with the most nominations). 

The axiomatic and approximation results highlight how flexibility in the number of prizes is crucial for designing impartial peer mechanisms with desirable properties. If the mechanism must always award a fixed number of prizes, discouraging manipulation can lead to undesirable outcomes, such as awarding the prize to a participant who does not receive any nominations. If the mechanism can award more prizes than planned or choose to leave the prize unassigned, manipulation can be discouraged while avoiding many of the undesirable outcomes that stem from a fixed number of prizes. The theoretical results also highlight the importance of randomization. In most cases, mechanisms that use randomization provide better approximation results than mechanisms that use a deterministic rule. 

Manipulation in peer mechanisms is not merely a theoretical concern. We survey empirical research that shows that people will try to manipulate peer mechanisms when given the chance. Employees will reduce the peer grades of coworkers when competing for promotion \citep{huang2019discovery}, and entrepreneurs bias peer reports in favor of friends and family when competing for loans \citep{hussam2021targeting}. Manipulation has also been shown in the lab. Whether experimental participants produce art \citep{balietti2016peer} or label envelopes \citep{carpenter2010}, they are happy to sabotage their peers to increase their own chance of winning a prize.

The empirical research provides several lessons for the theoretical study of peer mechanisms. First, participants often make errors in their peer evaluations. Mechanisms that are robust to errors are more likely to succeed in practice. Second, nepotism is common. Most models assume that participants care only about their own chance of winning a prize and not whether other perhaps related participants win a prize. Third, small amounts of manipulation may be acceptable. Rather than aiming to disincentivize manipulation from all participants, mechanisms could allow some manipulation and still yield good outcomes. Fourth, choosing the optimal manipulation can be difficult for participants. Complexity may be a useful tool to discourage manipulation.

Models of peer mechanisms differ from both the classic approach to mechanism design and the classic approach to social choice. The classic approach to mechanism design is to elicit information from individuals about themselves, such as the amount an individual would be willing to pay for an item at an auction. In peer mechanisms, individuals hold preferences or information about other participants (their peers). The classic approach to social choice is to aggregate the preferences of voters about a set of candidates. The voters and candidates are distinct. In peer mechanisms, the participants are both voters and candidates.

We focus this survey on preventing manipulation in peer mechanisms. We do not include research that focuses on how to aggregate nominations, rankings, or grades. Examples of this line of research include \citet{caragiannisgrading16,caragiannis2020}, which considers how counting methods can aggregate partial rankings, and \citet{wang2019}, which considers how to aggregate grades from individuals that have different standards and ranges. We do not include a recent line of research that studies how to incentivize participants to invite their peers to participate in a mechanism \citep{zhao2021}. We restrict our focus to settings where all participants are aware of the mechanism and the prize. We do not include research on peer grading that designs mechanisms to encourage peer graders to exert effort when grading (see \citet{zarkoob2023} for a recent example of this line of work). Our focus is on peer grading mechanisms that prevent graders from improving their own grades or rankings through manipulation. 

We include peer prediction mechanisms that have been adapted to evaluating information about people, such as a person's need for financial aid or their entrepreneurial ability. Typically, peer prediction is used for reports about external objects, such as the quality of a product or the forecast of an event. Not to be confused with the ``peer'' in peer mechanisms, the ``peer'' in peer prediction refers to the way these mechanisms use reports from multiple participants to incentivize truthful reports without access to ground truth to check the reports. Peer prediction mechanisms make payments to participants that depend on the participant's report and the reports of other participants who evaluate the same target object. We are interested in cases when the target object is information about another participant. See \citet{faltings2017peerprediction} for a survey of peer prediction mechanisms.

Our survey has some overlap with a recent survey of academic peer review by \citet{shah2022}, but our surveys make distinct contributions. Rather than focusing on a single application (academic peer review in the case of \citet{shah2022}), we focus on manipulation in peer mechanisms, which extends to several other applications, such as poverty targeting and peer grading of student assignments. \possessivecite{shah2022} survey covers some work on manipulation but does not go into the same detail as we do. Also, the models we discuss are often different. The models of peer mechanisms we discuss only correspond to models of academic peer review when each author submits one sole-authored paper and is also available as a reviewer. Authors often submit multiple papers, each paper can have multiple authors, authors may not act as reviewers, and reviewers may not submit papers.

We begin the survey with a motivating example to describe why many peer mechanisms create an opportunity for the participants to manipulate who wins the prize. We then provide a taxonomy of approaches to prevent manipulation and discuss the range of techniques that researchers have proposed. To highlight the two main theoretical approaches of axiomatic and approximation analysis, we focus on the model of peer selection. We survey the empirical studies of peer mechanisms and list key lessons the empirical research provides for theory. We conclude the survey by highlighting several areas in need of further research.


\section{Motivating Example}

Suppose a group of people compete for a prize by participating in a peer mechanism. The mechanism determines who wins the prize by asking each participant to nominate one or more peers and awarding the prize to the participant who receives the most nominations. Assume there is only one prize, which cannot be split between multiple participants. In the case of a tie, the prize is awarded uniformly at random between those with the most nominations.

One temptation is for each participant to nominate themselves, so we may want the mechanism to exclude self-nominations.%
\footnote{As \citet{ng2003} and \citet{ohseto2012} have shown, excluding self-evaluations can create problems in aggregating peer grades. Suppose participants $a$ and $b$ unanimously grade $a$ higher than $b$, but $a$ uses more generous grades than $b$. Excluding self-evaluations will discard $a$'s generous grade about himself but keep the generous grade he gives to $b$, which may cause $b$ to have a higher aggregated grade than $a$. \citet{ng2003} provides theoretical results highlighting an incompatibility between unanimity and excluding self-evaluations. \citet{ohseto2012} strengthen \possessivecite{ng2003} results to show that if participants can select grades from a finite and large set of real numbers, no aggregation rule excludes self-evaluations and satisfies monotonicity and unanimity.}
But even when the participants cannot nominate themselves, they may still have opportunities to manipulate who wins.

A simple example with three participants ($a$, $b$, and $c$) demonstrates that they can still manipulate who wins the prize. Suppose that: %
\begin{itemize}
    \item $a$ nominates $b$ and $c$,
    \item $b$ nominates $c$, and
    \item $c$ nominates $a$.
\end{itemize} %
Since both $b$ and $c$ have two nominations each, the mechanism awards the prize randomly to $b$ or $c$ with equal probability. Let's consider $c$'s perspective. All else equal, $c$ can ensure he wins the prize by nominating no one or nominating $a$. Even if $c$ believes that $b$ is worthy of the prize, $c$ has a strong incentive to manipulate the outcome to increase his chance of winning.

Whether the participants are asked to nominate, rank, or grade their peers, the same temptation remains. To increase their own chance of winning a prize, participants in a peer mechanism are tempted to manipulate their evaluation of their closest competitors. In the example, $c$'s closest competitor is $b$. By failing to nominate $b$, $c$ increases his chance of winning the prize at $b$'s expense.

\afterpage{
\begin{landscape}
\begin{table}
\vspace*{-2cm}
\thisfloatpagestyle{empty} 
\caption{Taxonomy of peer mechanisms}
\label{table:taxonomy}
\centerfloat
\small
 \begin{tabular}{m{70mm} m{40mm} m{28mm} m{28mm} m{16mm} m{34mm}}
 \toprule
  & \vbarmc{3}{Model} & \vbarmc{2}{Mechanism} \\ \cmidrule{2-6}
Paper   &	 Input	&	Output	&	Information & Approach	&	Technique \\
 \midrule
\citet{holzman2013impartial}	&	Nominate one peer	&	Single winner	&	Subjective	&	Impartial	&	Partition	\\
\citet{mackenzie2015symmetry}	&	Nominate one peer	&	Single winner	&	Subjective	&	Impartial	&	Random dictatorship	\\
\citet{babichenko2020forests}	&	Nominate one peer	&	Single winner	&	Subjective	&	Impartial	&	Expand possible winners	\\
\citet{edelman2021impartial}	&	Nominate one peer	&	Single winner	&	Subjective	&	Impartial	&	Random dictatorship	\\
\citet{cembrano2023improved} & Nominate one peer & Single winner & Subjective & Impartial & Permutation \\
\citet{amoros2011} & Nominate one peer & Single winner & Common & Impartial & Fix position \\
\citet{mackenzie2020pope} & Nominate one peer & At most one winner & Subjective & Impartial & Threshold \\
\citet{bjelde2017impartial}	&	Nominate one peer	&	At most two winners	&	Subjective	&	Impartial	&	Permutation	\\
\citet{tamura2014impartial,tamura2016characterizing}	&	Nominate one peer	&	Top $k$ winners	&	Subjective	&	Impartial	&	Expand possible winners	\\
\citet{fischer2014optimal,fischer2015optimal}	&	Nominate one or more peers	&	Single winner	&	Subjective	&	Impartial	&	Permutation	\\
\citet{bousquet2014near}	&	Nominate one or more peers	&	Single winner	&	Subjective	&	Impartial	&	Permutation	\\
\citet{babichenko2018diffusion}	&	Nominate one or more peers	&	Single winner	&	Subjective	&	Impartial	&	Expand possible winners	\\
\citet{caragiannis2019additive}	&	Nominate one or more peers	&	Single winner	&	Subjective	&	Impartial	&	Jury	\\
\citet{zhang2021}	&	Nominate one or more peers	&	Single winner	&	Subjective	&	Impartial	&	Expand possible winners	\\
\citet{caragiannis2021prior}	&	Nominate one or more peers	&	Single winner	&	Subjective	&	Impartial	&	Threshold	\\
\citet{cembrano2022additive}	&	Nominate one of more peers	&	Single winner	&	Subjective	&	Impartial	&	Threshold	\\
\citet{ito2018}	&	Nominate one or more peers	&	Single winner	&	Subjective	&	Reward	&	Peer prediction	\\
\citet{zhao2023} & Nominate one or more peers & Single winner & Subjective & Impartial & Expand possible winners \\
\citet{alon2011sum}	&	Nominate one or more peers	&	Top $k$ winners	&	Subjective	&	Impartial	&	Partition	\\
\citet{cembrano2022correspondences} &	Nominate one or more peers	&	Top $k$ winners	&	Subjective	&	Impartial	&	Expand possible winners	\\ 
\citet{was2019centrality} & Nominate one or more peers & Grade & Subjective & Impartial & Fix grade \\
\citet{li2018two}	&	Nominate top participant	&	Ranking	&	Common	&	Impartial	&	Fix position	\\
\citet{bao2021deterrence}	&	Nominate network neighbor	&	Single loser	&	Ground truth	&	Audit	&	Compare to nominee	\\
 \bottomrule
\end{tabular}
\end{table}

\begin{table}
\thisfloatpagestyle{empty} 
\caption*{Table 1 continued: Taxonomy of peer mechanisms}
\centerfloat
\small
 \begin{tabular}{m{70mm} m{40mm} m{28mm} m{28mm} m{16mm} m{34mm} }
 \toprule
  & \vbarmc{3}{Model} & \vbarmc{2}{Mechanism} \\ \cmidrule{2-6}
Paper   &	 Input	&	Output	&	Information & Approach	&	Technique \\
 \midrule
\citet{mattei2020peernomination,lev2021peer, lev2023}	&	Rank $m$ peers	&	Top $k$ winners	&	Common	&	Impartial	&	Threshold	\\
\citet{merrifield2009telescope}	&	Rank $m$ peers	&	Ranking	&	Common	&	Reward	&	Reward consensus	\\
\citet{xu2019conflictgraph}	&	Rank $m$ peers	&	Ranking	&	Subjective	&	Impartial	&	Partition	\\
\citet{stelmakh2020catch}	&	Rank $m$ peers	&	Ranking	&	Ground truth	&	Audit	&	Target manipulation	\\
\citet{bloch2021practice}	&	Rank network neighbors	&	Top $k$ winners	&	Ground truth	&	Impartial	&	Threshold	\\
\citet{bloch2020fbr}	&	Rank network neighbors	&	Ranking	&	Common	&	Impartial	&	Fix position	\\
\citet{amoros2002, amoros2023rank}	&	Rank all peers	&	Ranking	&	Common	&	Impartial	&	Fix position	\\
\citet{kahng2018ranking}	&	Rank all peers	&	Ranking	&	Subjective	&	Impartial	&	Partition	\\
\citet{berga2022impartial}	&	Rank all peers	&	Ranking	&	Subjective	&	Impartial	&	Partition	\\
\citet{cembrano2023ranking} & Rank all peers & Ranking & Subjective & Impartial & Fix position \\
\citet{hussam2021targeting}	&	Rank or grade $m$ peers	&	Single winner	&	Ground truth	&	Reward	&	Peer prediction	\\
\citet{rai2002targeting}	&	Binary type	&	Grades	&	Ground truth	&	Audit	&	Target disagreement	\\
\citet{declippel2008impartial}	&	Relative grade	&	Cardinal share	&	Subjective	&	Impartial	&	Reduce total reward	\\
\citet{kurokawa2015impartial}	&	Grade $m$ peers	&	At most $k$ winners	&	Subjective	&	Impartial	&	Expand possible winners	\\
\citet{aziz2016strategyproof,aziz2019strategyproof}	&	Grade $m$ peers	&	Top $k$ winners	&	Subjective	&	Impartial	&	Partition	\\
\citet{dhull2022}	&	Grade $m$ peers	&	Top $k$ winners	&	Subjective	&	Impartial	&	Partition	\\
\citet{wang2018tsp}	&	Grade $m$ peers	&	Top $k$ winners	&	Common	&	Impartial	&	Partition	\\
\citet{chakraborty2018}	&	Grade $m$ peers	&	Grades	&	Ground truth	&	Audit	&	Assign audited peers	\\
\citet{dealfaro2014crowdgrader}	&	Grade $m$ peers	&	Grades	&	Ground truth	&	Reward	&	Reward consensus	\\
\citet{baumann2018self}	&	Grade network neighbors	&	Single winner	&	Ground truth	&	Audit	&	Limit misreports	\\
\citet{babichenko2020network}	&	Grade network neighbors	&	Top $k$ winners	&	Subjective	&	Impartial	&	Expand possible winners	\\
\citet{cembrano2023ranking} & Grade all peers & Top $k$ winners	& Subjective & Impartial & Partition \\
\citet{walsh2014peerrank}	&	Grade all peers	&	Grades	&	Ground truth	&	Reward	&	Reward consensus	\\
\citet{niemeyer2022}	&	Abstract message space	&	Single winner	&	Subjective	&	Impartial	&	Jury	\\
 \bottomrule
\end{tabular}
\end{table}

\end{landscape}
}


\section{A Taxonomy of Models}\label{sec:taxonomy}

We surveyed the literature for peer mechanisms that address manipulation. Since peer mechanisms are inspired by a variety of applications, there are a variety of different models to describe these applications. In Table \ref{table:taxonomy}, we provide a taxonomy of the models we uncovered. We distinguish between different models according to their:
\begin{itemize}[leftmargin=8em,itemsep=0.2em]
    \item[Input:]  What do the participants need to report about their peers?
    \item[Output:]  What output does the mechanism produce?
    \item[Information:]  What type of information do participants hold about their peers?
\end{itemize} %
We have some notation within Table \ref{table:taxonomy}. We use $m$ for the number of peers each participant is asked to evaluate. In most cases, $m$ is small relative to the number of participants. We use $k$ for the number of winners when the mechanism selects multiple winners.

The table also includes columns to categorize the mechanisms according to:
\begin{itemize}[leftmargin=8em,itemsep=0.2em]
    \item[Approach:]  Does the mechanism use audits, rewards, or impartiality to discourage manipulation?
    \item[Technique:]  What technique does the mechanism use?
\end{itemize}

The taxonomy is useful for several reasons. First, the contributions to peer mechanisms are spread across computer science and economics, and the taxonomy shows which contributions are most closely connected. Second, the taxonomy shows gaps in the literature. For example, the ``Approach'' column of Table \ref{table:taxonomy} shows that most of the existing work focuses on constructing impartial mechanisms. Less work has focused on using audits or rewards to discourage manipulation.

\subsection{Inputs}

The input into a peer mechanism is the reports from the participants about their peers. Models differ by the detail of the reports and which peers they can report on. In increasing order of detail, the mechanism could ask for a nomination, a ranking, or a grade. A nomination asks the participant to select one or more peers. A ranking asks for a strict order of peers. A grade asks for a cardinal score for each peer.

The appropriate form of peer report can be linked to the level of information each participant holds about their peers. If participants are only able to sort their peers into two groups, such as worthy and unworthy for the prize, nominations are appropriate. If they have more detailed information to order peers but not enough to determine a cardinal score, a ranking is appropriate. Finally, the most detailed information can be modeled by a cardinal score.

To our knowledge, there is little research on the form of peer report that participants prefer. In the context of peer grading, \citet{dealfaro2014crowdgrader} reported that:
\begin{quote}
``Students expressed some uneasiness in ranking their peers, especially as they perceived ranking as a blunt tool, unable to capture the difference between a pair of roughly equivalent submissions, and a pair of submissions, one of which was very good, and the other non-functional.''
\end{quote}
Further empirical research will be needed to guide theory on the appropriate form of peer reports in peer grading and other contexts. 

Although most models use either peer nominations, ranks, or grades, there are some exceptions. \citet{rai2002targeting} uses a model with two participants that can be a binary type (either rich or poor). Each participant reports whether they are poor and whether their peer is poor. However, \possessivecite{rai2002targeting} model can be thought of as a model of nominations where self-nominations are allowed. Participants can choose to nominate no one, nominate themselves only, their peer only, or both themselves and their peer. Another exception is \citet{niemeyer2022}, who use an abstract message space. They do not restrict how the participants can communicate with the mechanism designer.

The mechanism must also specify which peers each participant can evaluate. Some models do not restrict which peers each participant can evaluate. Each participant considers all his peers in his reports. Others assume that each participant reports on a fixed number of peers (represented by $m$ in Table \ref{table:taxonomy}). When the size of the community is large, reporting on every peer may be too onerous.\footnote{The burden of grading all peers can be reduced by combining nominations and grades. The mechanism in \citet{cembrano2023weights} asks each participant to nominate as many peers as they would like and assign grades (or weights) to each nomination. The peers that are not nominated receive a grade of zero.} Distributing peer reports equally among participants is a natural alternative.

Another class of models assumes that the participants are connected by a network, and each participant can only evaluate his network neighbors. For example, in the network shown below, $a$ can evaluate $b$ and $c$, but cannot evaluate $d$ and $e$. 

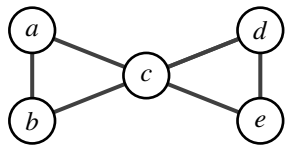
\begin{center}
 \begin{tikzpicture}
\Vertex[label=$c$,fontsize=\normalsize,color=white]{c} 
\Vertex[x=-1.5,y=0.6,label=$a$,fontsize=\normalsize,color=white]{a}
\Vertex[x=-1.5,y=-0.6,label=$b$,fontsize=\normalsize,color=white]{b}
\Vertex[x=1.5,y=0.6,label=$d$,fontsize=\normalsize,color=white]{d}
\Vertex[x=1.5,y=-0.6,label=$e$,fontsize=\normalsize,color=white]{e}
\Edge(c)(b)
\Edge(c)(a)
\Edge(c)(d)
\Edge(c)(e)
\Edge(c)(d)
\Edge(b)(a)
\Edge(d)(e)
\end{tikzpicture}   
\end{center}

The network can capture situations where participants may learn about their peers through social interactions. The network could represent friendships or co-worker relationships. Social networks often display common structures, such as clustering---if $a$ is friends with $b$ and $c$, then $b$ and $c$ are likely to be friends. A recent line of research, initiated by \citet{baumann2018self} and \citet{bloch2020fbr} in economics and \citet{babichenko2020network} in computer science, investigates how the structure of the network impacts the designer's ability to construct mechanisms that perform well. 

The network can also represent who the participants are not allowed to report on. For example, in the context of conference peer review, the network could represent conflicts of interest, such as co-author and student-supervisor relationships \citep{xu2019conflictgraph}. The network of conflicts is the complement or inverse of the network studied by \citet{baumann2018self}, \citet{bloch2020fbr} and \citet{babichenko2020network}.

Once the network is defined, the model must still define the form of the peer reports. Participants may be asked to nominate \citep{bao2021deterrence}, rank \citep{bloch2020fbr}, or grade \citep{baumann2018self,babichenko2020network} their network neighbors. 

In \citet{baumann2018self}, \citet{bloch2020fbr} and \citet{babichenko2020network}, the network that captures the ability to evaluate peers is unweighted.\footnote{After the participants grade their peers in \possessivecite{babichenko2020network} model, the grades can be represented as a weighted network.} Either the participant can evaluate a given peer or they cannot. \citet{dhull2022} uses a weighted network where the weight measures expertise to evaluate a given peer. The higher the weight, the more accurate the peer evaluation. In the context of conference peer review, the weight can be thought of as a similarity between two authors' research interests. \citet{dhull2022} study how to assign paper submissions to evaluators to maximize the accuracy of evaluations while ensuring that evaluators' own submissions are not competing with the submissions they are called to evaluate. (This is done using the partition mechanism, which we discuss in more detail in Section \ref{subsection:impartial}.)

\subsection{Outputs}\label{sec:outputs}

The output of a peer mechanism describes the form of the prize. Models differ in the number of prizes and if the prizes differ in quality. Inspired by best paper awards at conferences or selecting the most influential user in a social media network, some models assume the output is a single winner. A participant's utility is defined by the probability he wins the prize. Other applications, such as research grants, have inspired models with multiple winners for a prize of equal quality. In Table \ref{table:taxonomy}, the number of winners is given by a constant $k$. Finally, peer grading and targeting of aid or loans have inspired models where the mechanism assigns a rank or a grade to each participant, and utility increases in the rank or grade. Similar to a grade, \citet{declippel2008impartial} defined the output as the share of a divisible prize.

A theme in the theoretical analysis of peer mechanisms (which we will expand on in Section \ref{sec:theory_results}) is that flexibility in the output of the mechanism allows the designer to construct mechanisms with desirable properties. For example, suppose a designer who aims to award a single prize has the flexibility to award a second prize to a runner-up. If the runner-up is close enough to the winner that the runner-up can divert the prize to himself by changing his peer report, the mechanism can simply award a second prize to the runner-up to discourage the temptation for the runner-up to manipulate who wins. 

Nearly every paper listed in Table \ref{table:taxonomy} assumes the prize is desirable. Participants want to be selected or improve their rank or grade. The only exception is a model of criminal networks studied by \citet{bao2021deterrence} where the selected participant pays a fine. Rather than selecting a single winner, the mechanism selects a single loser.

\subsection{Information}

Peer mechanisms focus on eliciting the information participants hold about their peers. How is the peer information generated? We divide the models into three categories:
\begin{itemize}
  \item Subjective information: an opinion about who is worthy of a prize, who should receive a higher rank or a higher grade.
  \item Common information: all participants agree about who is worthy or who should receive a higher rank or grade.
  \item Ground truth information: each participant has a value which can be checked through an audit or some other measure. 
\end{itemize}

The types of peer information in the above list are ordered from least to most constrained. Subjective information allows the participants to hold any opinions about their peers. Common information constrains all peers to agree. Ground truth information adds an additional requirement to common information that the information can be checked by the mechanism designer.

A model with common information or ground truth information does not imply that all participants have identical information. Each participant may only hold information about a subset of peers. One participant may know that $a$ is ranked above $b$ while another may know that $b$ is ranked above $c$ but have no knowledge about $a$. Also, the participants may make errors if the common or ground truth information is observed with noise. For example, \citet{lev2021peer} use the Mallows model to shuffle the ranking that each participant observes. The Mallows model has a dispersion parameter that ranges from perfect information at 0 to no information at 1. As the parameter approaches 0, participants observe a ranking that is concentrated around the true common ranking. At 1, each participant draws a ranking uniformly at random from all possible rankings. 

Some participants may make systematically more errors than others. In the context of peer grading, students who have a good understanding of an assignment may grade their peers more accurately than students who struggle to understand the assignment. The mechanism designer may wish to give more weight to graders who got good grades themselves \citep{walsh2014peerrank}. Grader accuracy or reliability may also be measured directly if the mechanism designer can compare a peer grade to ground truth, such as the grade given by an instructor \citep{chakraborty2018}.

Each type of information matches different applications. Social media (such as Twitter) is an example of an application with subjective information. Each user follows peers who they find interesting, but users may disagree on which peers are interesting. The disagreement does not imply that one user is wrong. Peer grading is an example of ground truth information. Each student has a ground truth score according to the marking rubric that could be verified by an expert grader. Common information has similar applications to ground truth, except that the model does not specify how the mechanism designer can verify the information. In the example of targeting aid, participants may agree which peers are most in need of aid, but the designer may not have a method to measure need. 

Almost all of the papers in Table \ref{table:taxonomy} assume the mechanism designer has no prior information on the peer reports. \citet{caragiannis2021prior} is the one exception. Their model assumes that the designer knows the probability that each participant will nominate each peer. The prior information is useful to select a default winner in the case of ties, as ties can create opportunities to manipulate the mechanism.

\subsection{Mechanisms: Approaches and Techniques}\label{sec:mechanisms}

We provide an overview of the three main approaches (audits, rewards, and impartial mechanisms) used to encourage truthfulness and discourage the manipulation of peer mechanisms. We also highlight the range of techniques used to implement each approach. The three approaches are not mutually exclusive. A single peer mechanism could use all three.

\subsubsection{Audits}

In some settings, an auditor can check the peer reports. For example, in peer grading of large courses, the instructor can check if students have graded their peers accurately, while in poverty targeting, the grant agency could conduct surveys to measure the poverty level of some households.

If audits can uncover the truth, why not audit everyone? In many applications, including peer grading and poverty targeting, we can assume that audits are more costly than peer reports. The goal is to undertake a limited number of audits to achieve the required performance.

One technique is to \emph{target disagreement} in peer reports \citep{rai2002targeting}. In the poverty targeting setting, if I claim I am poor, but my neighbor says I am rich, the mechanism can audit the disagreement and punish misreports. The mere possibility of an audit can discourage misreporting and may produce a desirable equilibrium where all participants report truthfully, and no audits need to be conducted.

Targeting disagreement is effective when the designer knows that the participants have perfect information about each other. When there is some error or noise in the peer reports, detecting manipulation becomes more difficult. Disagreement no longer implies that one of the participants is lying. Participants could hold different information about their peers by chance and disagree even when they are both reporting truthfully. For the designer to \emph{target manipulation}, they need to determine if a participant's peer report improves his position by chance or due to manipulation \citep{stelmakh2020catch}.

Audits may also limit the extent to which participants can lie about their peers. When employees compete for a promotion, they may need to support claims about peers and themselves with some evidence. The mechanism can exploit the need for evidence \emph{limits misreports} about themselves and their peers \citep{baumann2018self}. The best employee can make the highest claim about his own performance, and the most negative peer review he can receive will be better than the most negative peer review any of his participants can receive.

In the constant of peer grading, the mechanism can audit a small number of student assignments and then \emph{assign the audited peers} assignments to all students for grading \citep{chakraborty2018}. Since the students do not know which assignments have been audited, the mechanism discourages manipulation for all assignments. 

The mechanism can use peer nominations to decide who to audit. In a model of criminal networks, the mechanism samples one participant and asks that participant to nominate one peer. The \emph{compare to nominee} mechanism audits both the sampled participant and the nominated peer but only fines the one with the higher level of criminal activity. Since audits only provide a signal of criminal activity, the sampled participant minimizes their chance of incurring a fine by nominating the peer who they think has the highest level of criminal activity. 

\subsubsection{Rewards}

To prevent manipulation, the mechanism designer can reward truthful reporting. The rewards can take the form of monetary payments or some other form. In peer grading, the reward might not be money but the student's grade. For instance, the mechanism in \citet{walsh2014peerrank} increases a student's grade when they grade other students well. 

If the peer reports are entirely subjective, then paying for truthful reports is not particularly applicable. A participant can always claim that their report is their true but subjective opinion. In Table \ref{table:taxonomy}, most of the mechanisms that use payments are in settings where there is common information or ground truth. The only exception is \citet{ito2018}, where each participant's nomination is subjective. However, the reward component of \possessivecite{ito2018} mechanism is used on a part of the model that does have common information---whether both participant $a$ and participant $b$ observe a nomination from $a$ to $b$.

If each peer has common information that all participants agree upon, a natural approach is to \emph{reward consensus}. Multiple peer assessments of the peer's value should converge. A participant who wishes to manipulate the mechanism must consider the cost of diverging from the consensus and reducing their reward. For example, if the prize is the ranking of a grant application, the participants can receive an increase in the ranking as payment for a report that agrees with the consensus \citep{merrifield2009telescope}.

Rewarding consensus can, however, create problems. If a participant believes that the consensus will be biased, he may bias his peer reports to match the consensus. The mechanism must give the participants an incentive to report their true assessment, even if they believe that their true assessment will differ from the consensus.

The field of \emph{peer prediction} uses payments to extract true assessments even though the designer does not have access to the ground truth \citep{faltings2017peerprediction}. In peer prediction, the focus is usually on assessing objects that are unrelated to the peers themselves, such as the quality of a product or a forecast of an event. The peer mechanisms we discuss in this survey involve peers assessing each other.

Peer prediction mechanisms can be adapted to peers assessing each other. For example, \citet{hussam2021targeting} adapt \possessivecite{witkowski2012rbts} peer prediction mechanism to peer reports. The mechanism asks each participant for peer reports and for their belief about what other participants' peer reports will be. If the other participants report truthfully, a participant maximizes their expected payment by truthfully reporting their peer report and their belief of other participants' peer reports.

\subsubsection{Impartial mechanisms}\label{subsection:impartial}

The bulk of the research surveyed in Table \ref{table:taxonomy} discourages manipulation by designing an impartial mechanism. A peer mechanism is impartial when a participant cannot influence his chances of receiving a prize or improving his rank or grade.\footnote{The definition of an impartial mechanism is slightly different from a strategy-proof mechanism. In a strategy-proof mechanism, an agent has a weakly dominant strategy to report the truth. As \citet{fischer2014optimal,fischer2015optimal} point out, impartiality is equivalent to strategy-proofness if the utility of the participant only depends on their chance of winning the prize. Strategy-proof mechanisms are also referred to as dominant-strategy incentive-compatible mechanisms.}  

Although most papers define impartiality so that a mechanism is impartial when every participant cannot change their own probability of receiving a prize or improve their own rank or grade, there are some exceptions. For example, \citet{berga2022impartial} use a stricter definition of impartiality for the case where the output of the mechanism is a ranking. In \possessivecite{berga2022impartial} definition, a mechanism is impartial when a participant cannot impact their own position in the ranking and who is below and who is above them in the ranking. In contrast, \citet{kahng2018ranking} and \citet{cembrano2023ranking} use the more standard definition of impartiality that each participant cannot change their own position in the ranking.

One technique to achieve impartiality is to \emph{partition} participants into groups \citep{alon2011sum,holzman2013impartial}. For example, the mechanism designer divides participants into two groups, $A$ and $B$. Participants in $B$ pick a winner in $A$, while participants in $A$ pick a winner in $B$. The overall winner is decided with a coin toss. Each individual can only influence the chance a peer outside of his or her group wins the prize, so the mechanism is impartial. 

How many participants should be in each group? At one extreme, the \emph{random dictatorship} randomly chooses a single participant for one group and places the rest of the participants in the other group. The single participant (the dictator) decides who wins the prize. \emph{Jury} mechanisms increase the number of participants in the dictatorship group, and these participants act as a jury to decide on the winner among the remaining participants. 

Partition mechanisms can have poor performance if all the best participants end up in the same group. For example, suppose a partition mechanism for selecting two winners divides the participants into two groups and selects one winner from the first group and one from the second group. If the top two participants end up in the first group, the mechanism will only select one of them. \citet{aziz2016strategyproof,aziz2019strategyproof} counter this problem by using more than two groups and deciding on the number of winners in each group based on the grades from peers outside the group.

One participant can also be part of many different partitions. In the \emph{permutation} mechanism introduced by \citet{fischer2014optimal,fischer2015optimal}, the participants are placed in random order, and the mechanism only counts nominations from peers that are before the participant in the order. To decide on the winner, the mechanism starts with the first participant as the candidate winner and moves through the order to update the candidate winner. A participant $p$ becomes the candidate winner if he is above the current candidate $c$ in the order, and the number of nominations $p$ receives from peers before $p$ in the order (excluding $c$) is greater than or equal to the number of nominations $c$ receives from peers before $c$ in the order. The winner is the candidate winner after moving through all participants in the order. The mechanism is impartial because each participant can only influence if a peer wins when the participant is no longer able to win. 

Researchers often assume the designer must award a fixed number of prizes, which creates a strong incentive for participants to misreport. For example, if the mechanism must award a single prize, the participant in second place has a strong incentive to share a negative review of the participant in first place.

If the designer has some flexibility in awarding the prize, impartiality can be achieved without resorting to a partition. One approach is to \emph{expand the set of possible winners} to include participants that could win if they changed their peer reports \citep{tamura2014impartial,kurokawa2015impartial}. If the designer is constrained to choose at most $k$ winners, she can randomly pick $k$ from the expanded set of possible winners. But, the probability that each participant is selected cannot depend on their peer report. One solution is to have the option of not selecting any winners to remove the participant's incentive to reduce the number of possible winners \citep{kurokawa2015impartial}. If the designer has no constraints on the number of winners, another approach is to choose an exogenous
\emph{threshold} and only award prizes to participants who exceed the threshold \citep{mattei2020peernomination}. By tweaking the mechanism, the designer can award $k$ prizes in expectation.

Some flexibility in the number of winners also helps to generalize the permutation mechanism described above. The permutation mechanism selects one winner. If the designer aims to select two winners, \citet{bjelde2017impartial} show that the permutation mechanism can be adapted if the mechanism is allowed to select one winner in a certain case. The permutation mechanism is run both forward and backward, and one winner is selected on each run. If the same participant is selected in both the forward and backward run, the mechanism only selects one winner.

If the prize is divisible, the designer can \emph{reduce the total reward} to achieve impartiality. Suppose the designer would like to share the prize according to the participants' reports when the participants have consensus. The designer can discourage deviations from consensus by reducing the share of the prize awarded to participants that disagree and increasing the share of a default participant \citep{declippel2008impartial}. 

Most of the research uses a strict definition of impartiality---a participant cannot influence whether he receives the prize no matter what his peers report. We could relax this definition so that the participant cannot influence whether he receives the prize provided his peers report truthfully. Rather than looking for strategy-proof or dominant-strategy incentive-compatible mechanisms, we look for ex post incentive-compatible mechanisms. Ex post incentive compatibility is achieved if each participant has no incentive to lie when all other participants report truthfully. Several papers, including \citet{amoros2002,amoros2011, amoros2023rank}, \citet{li2018two}, \citet{bloch2020fbr} and \citet{babichenko2020forests}, use more relaxed definitions of impartiality and incentive-compatibility to construct mechanisms with good performance.

If the mechanism only needs to meet ex post incentive compatibility, the designer can construct what we refer to as \emph{fix position} mechanisms. Provided all other participants report the truth, the participant's position in the output is fixed, and he cannot change his probability of winning the prize. Consider the following example of a fix position mechanism. Suppose three participants, $a$, $b$, and $c$, are asked to rank their peers to win a prize, and the true order is $a$ first, $b$ second, and $c$ last. If all participants report truthfully, do they have an incentive to deviate from this equilibrium? Consider $b$'s perspective when both $a$ and $c$ have reported truthfully that $a \succ b$ and $b \succ c$. No matter what $b$ reports, the designer can still use $a$ and $c$'s reports to fix $b$ in second place. At equilibrium, $b$'s report does not influence his chance of winning the prize.

The fix position mechanism applies when the mechanism's output is a ranking or selection of participants. When the mechanism outputs a grade, and the participants care only about their grade (and not their relative position), a simple way to achieve impartiality is to \emph{fix the grade} of each participant to depend only on other participants' peer reports \citep{was2019centrality}. Participants can change other participants' grades but not their own.

We conclude this section with the disclaimer that impartiality, alone, does not guarantee a good outcome. If each participant cares only whether they receive a prize and the mechanism is impartial, each participant will be indifferent between reporting their true knowledge or preferences about their peers and any other report. This indifference gives rise to multiple equilibria, some of which may have undesirable outcomes.\footnote{Some models, such as \citet{amoros2002}, assume that participants prefer to report truthfully if their report does not influence their own chances of winning a prize. This assumption avoids many undesirable equilibria.} For example, suppose an impartial mechanism selects a winner who would not receive any nominations under the participants true preferences. The participants have no incentive to change their report even though their change may cause a more desirable outcome. Mechanisms need to satisfy other properties in addition to impartiality to guarantee desirable outcomes. In the next section, we discuss the other properties that impartial mechanisms can satisfy.


\section{Theoretical Results} \label{sec:theory_results}

The theoretical results on peer mechanisms illuminate the limits and potential of peer mechanisms. Researchers have taken two main approaches: proving which mechanisms (if any) satisfy a set of axioms and showing how certain impartial mechanisms can approximate the optimal truthful outcome.

To highlight the differences between the two approaches, we focus on peer selection, where each participant can nominate peers to receive a prize. As Table \ref{table:taxonomy} shows, the model of peer selection is popular. Nominations are used as the input of the mechanism in 21 of the 37 papers on impartial peer mechanisms. 

The axiomatic study of peer selection was initiated by \citet{holzman2013impartial} while the approximation approach was initiated by \citet{alon2011sum}. In the following two subsections, we describe results from each of these two seminal works and the papers that built upon them. 

The nominations can be represented as a directed graph. An edge from a participant to a peer shows that the participant nominates that peer. In the graph below, $a$ nominates $b$ and $b$ nominates $c$. Participant $a$ is nominated by $c$, $d$, and $e$ so $a$ receives the most nominations.

\begin{center}
\begin{tikzpicture}
    \Vertex[x=-0.809,y=0.588,label=$a$,fontsize=\normalsize,color=white]{a}
    \Vertex[x=0.309,,y=0.951,label=$b$,fontsize=\normalsize,color=white]{b}
    \Vertex[x=1,y=0,label=$c$,fontsize=\normalsize,color=white]{c}
    \Vertex[x=0.309,y=-0.951,label=$d$,fontsize=\normalsize,color=white]{d}
    \Vertex[x=-0.809,y=-0.588,label=$e$,fontsize=\normalsize,color=white]{e}
\Edge[Direct](a)(b)
\Edge[Direct](b)(c)
\Edge[Direct](c)(a)
\Edge[Direct](d)(a)
\Edge[Direct](e)(a)
\end{tikzpicture}
\end{center}

\subsection{Axioms}

\citet{holzman2013impartial} use a model where each participant can nominate one peer, and there is a single prize. Self-nominations are not allowed. In this model, \citet{holzman2013impartial} show that impartial peer mechanisms fail to satisfy two weak axioms simultaneously:
\begin{description}
\item[positive unanimity:] A participant always wins if he is nominated by everyone else.
\item[negative unanimity:] The winner gets at least one nomination.
\end{description}

\begin{mytheorem}[\citealp{holzman2013impartial}]
There exists no nomination rule that satisfies impartiality, positive unanimity, and negative unanimity
\end{mytheorem}

The strong impossibility result raises the question of whether similar results apply in different (perhaps more complex) models. If we allow the mechanism to award more than one prize, the impossibility no longer holds. \citet{tamura2014impartial} show that a peer mechanism with single nominations and more than one prize can simultaneously satisfy impartiality, positive unanimity, and negative unanimity, provided there are at least four participants. 

To prove the result, \citet{tamura2014impartial} defines a mechanism called ``plurality with runners-up''. The participant with the most nominations wins a prize. If there is a tie for the most nominations, all the tied participants win a prize. Each runner-up also wins if and only if he nominates the single participant with the most nominations who only wins by one point.

The downside of plurality with runners-up is that every participant could win a prize, and, in this situation, a single participant could reduce the number of winners from everyone to just himself and one other by changing his nomination. Consider the example below; there is a cycle of nominations. Since every participant receives one nomination, plurality with runners-up awards a prize to everyone.

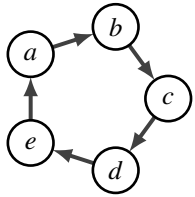
\begin{center}
\begin{tikzpicture}
    \Vertex[x=-0.809,y=0.588,label=$a$,fontsize=\normalsize,color=white]{a}
    \Vertex[x=0.309,,y=0.951,label=$b$,fontsize=\normalsize,color=white]{b}
    \Vertex[x=1,y=0,label=$c$,fontsize=\normalsize,color=white]{c}
    \Vertex[x=0.309,y=-0.951,label=$d$,fontsize=\normalsize,color=white]{d}
    \Vertex[x=-0.809,y=-0.588,label=$e$,fontsize=\normalsize,color=white]{e}
\Edge[Direct](a)(b)
\Edge[Direct](b)(c)
\Edge[Direct](c)(d)
\Edge[Direct](d)(e)
\Edge[Direct](e)(a)
\end{tikzpicture}
\end{center}

Suppose $a$ switched his nomination from $b$ to $c$ (as shown below). Since $c$ receives the most nominations, he would still win a prize. Participants $a$, $b$, $d$, and $e$ are all one nomination behind $c$, but only $a$ would continue to win a prize. Recall that plurality with runners-up only awards a prize to a runner-up if he nominates the single participant with the most nominations who only wins by one nomination. 

\begin{center}
\begin{tikzpicture}
    \Vertex[x=-0.809,y=0.588,label=$a$,fontsize=\normalsize,color=white]{a}
    \Vertex[x=0.309,,y=0.951,label=$b$,fontsize=\normalsize,color=white]{b}
    \Vertex[x=1,y=0,label=$c$,fontsize=\normalsize,color=white]{c}
    \Vertex[x=0.309,y=-0.951,label=$d$,fontsize=\normalsize,color=white]{d}
    \Vertex[x=-0.809,y=-0.588,label=$e$,fontsize=\normalsize,color=white]{e}
\Edge[Direct](a)(c)
\Edge[Direct](b)(c)
\Edge[Direct](c)(d)
\Edge[Direct](d)(e)
\Edge[Direct](e)(a)
\end{tikzpicture}
\end{center}

\citet{tamura2014impartial} show that it is possible to adjust plurality with runners-up to have at most two winners while simultaneously satisfying impartiality, positive unanimity, and negative unanimity. However, the adjustment requires that in the case of a tie for the most nominations, the participant who is earlier in a pre-defined order always wins. Thus, participants who are earlier in the order have an advantage over participants who are later in the order. 

Favoring certain applicants who happen to be earlier in an order may be an undesirable property. To address this challenge, \citet{tamura2016characterizing} search for impartial peer mechanisms that satisfy the following axioms:
\begin{description}
  \item[Symmetry:] The determination of the winners is independent of the order of the participants.
  \item[Anonymity:] An exchange of nominations between two participants does not affect whether any other participant wins.
  \item[Monotonicity:] Receiving an additional nomination cannot cause a winner to lose his prize.
\end{description}

\begin{mytheorem}[\citealp{tamura2016characterizing}]
Plurality with runners-up is the only minimal nomination rule satisfying impartiality, symmetry, anonymity, and monotonicity.
\end{mytheorem}

The definition of \emph{minimal} is that the nomination rule must have the smallest set of winners while still satisfying the four axioms. For example, a rule that always gives a prize to every participant would satisfy impartiality, symmetry, anonymity, and monotonicity but would not be minimal since plurality with runners-up can satisfy the same axioms and choose a strictly smaller number of winners for at least one profile of nominations.

Anonymity is a desirable property because it gives the participants privacy when they report their nominations. The winner can be determined even if the participants complete their nominations anonymously. Unfortunately, anonymity is a difficult property for impartial mechanisms to satisfy. 

If we return to the single winner setting of \citet{holzman2013impartial}, the only impartial nomination rules that satisfy anonymity give the prize to the same default participant---no matter the profile of nominations. However, the result only applies to deterministic mechanisms. If the mechanism designer is willing to use randomization, \citet{mackenzie2015symmetry} shows that:

\begin{mytheorem}[\citealp{mackenzie2015symmetry}]
An impartial nomination rule satisfies anonymity and negative unanimity if and only if it is a uniform random dictatorship.
\end{mytheorem}

The uniform random dictatorship picks a single participant uniformly at random, and this participant picks the winner. If the nominations are placed anonymously in a box, the uniform random dictatorship can be implemented by randomly selecting one of the nominations.\footnote{Anonymity and negative unanimity is not the only way to characterize the uniform random dictatorship. \citet{edelman2021impartial} show that the rule can also be characterized by other axioms.} 

Besides randomization, are there other features of the model that can allow the designer to break away from the incompatibility between impartiality and anonymity? \possessivecite{holzman2013impartial} model assumes that a winner must be selected---the prize cannot remain unassigned. If the prize can remain unassigned, a threshold mechanism can satisfy impartiality, anonymity, and other desirable axioms, such as positive unanimity and monotonicity. The threshold mechanism assigns the prize to the participant who receives at least a threshold number of nominations and leaves the prize unassigned otherwise. The threshold must be greater than half the number of nominations to ensure that there is, at most, a single winner. \citet{mackenzie2020pope} shows that the threshold mechanism is the only deterministic impartial nomination rule that satisfies anonymity, monotonicity, positive unanimity, and candidate neutrality (all participants are treated symmetrically when they are considered as possible winners).

The results on anonymity we have discussed thus far use nominations as the input to the mechanism. Do other inputs, such as ranks or grades, also display a tension between impartiality and anonymity? \citet{berga2022impartial} show that when participants are asked to rank all their peers and the mechanism outputs a ranking, impartiality and anonymity are incompatible. Similar to \possessivecite{holzman2013impartial} result for nominations, the only mechanisms satisfying impartiality and anonymity for ranking output the same default ranking---ignoring the participants' reports.

\subsection{Approximation}

If everyone reported truthfully, the ideal mechanism would be to award the prize to the participant who receives the most nominations. Unfortunately, this ideal mechanism is not impartial. For example, suppose $a$ and $b$ both receive the same number of nominations, and one of the nominations $b$ receives is from $a$. By simply changing his nomination to any other peer, $a$ can reduce the number of nominations $b$ receives and put himself in the top position.

Can we design a mechanism that is close to the ideal but still retains impartiality? A line of research initiated by \citet{alon2011sum} seeks to answer this question by designing mechanisms that are impartial and closely approximate the ideal of awarding the prize to the participant with the most nominations. \citet{alon2011sum} defined the \emph{approximation ratio} as the number of nominations the selected participant receives divided by the number of nominations received by the participant with the most nominations. The closer this ratio is to $1$, the better the mechanism performs. 

To better understand the approximation ratio, consider the example shown below. Participant $a$ receives 3 nominations, which is the most nominations received by any participant. Participants $b$ and $c$ both receive 1 nomination and $d$ and $e$ receive zero nominations. If the mechanism selected participant $b$, the approximation ratio is $\frac{1}{3}$.   

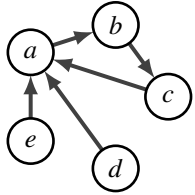
\begin{center}
\begin{tikzpicture}
    \Vertex[x=-0.809,y=0.588,label=$a$,fontsize=\normalsize,color=white]{a}
    \Vertex[x=0.309,,y=0.951,label=$b$,fontsize=\normalsize,color=white]{b}
    \Vertex[x=1,y=0,label=$c$,fontsize=\normalsize,color=white]{c}
    \Vertex[x=0.309,y=-0.951,label=$d$,fontsize=\normalsize,color=white]{d}
    \Vertex[x=-0.809,y=-0.588,label=$e$,fontsize=\normalsize,color=white]{e}
\Edge[Direct](a)(b)
\Edge[Direct](b)(c)
\Edge[Direct](c)(a)
\Edge[Direct](d)(a)
\Edge[Direct](e)(a)
\end{tikzpicture}
\end{center}

If the designer is restricted to deterministic mechanisms, the results are disappointing. \citet{alon2011sum} find that no deterministic mechanism can provide a finite approximation ratio. The result mirrors \possessivecite{holzman2013impartial} result that an impartial deterministic mechanism may assign the prize to a participant who does not receive any nominations.

Randomization provides more encouraging results. \citet{alon2011sum} show that the partition mechanism that divides the participants into two equal-size groups, only counts nominations across groups, and randomly picks which group the winner is selected from, provides an approximation ratio of $\frac{1}{4}$ in expectation. Since only the between-group nominations are counted, each nomination is counted with probability $\frac{1}{2}$. And the winner is selected from a given group with probability $\frac{1}{2}$.

In the partition mechanism with two groups, the mechanism may ignore many of the nominations. To attain better approximation ratios, \citet{fischer2014optimal,fischer2015optimal} tackle this problem by increasing the number of partitions and allowing participants to be part of many different partitions. Since each participant is part of many different partitions, they call their mechanism the ``permutation mechanism". The permutation mechanism achieves an approximation ratio of $\frac{1}{2}$, which turns out to be the best possible approximation ratio.

Consider the example shown below on the left. The two participants, $a$ and $b$, both nominate each other. Without loss of generality, let's focus on participant $a$. To achieve impartiality, the probability that $a$ wins must be the same whether they nominate $b$ or abstain (the situation shown below on the right). Thus, the probability $a$ wins must be $\frac{1}{2}$ when they abstain and $b$ nominates them. But now $b$ must win with probability $\frac{1}{2}$ even though they do not receive any nominations. This situation shows that any mechanism cannot be more than $\frac{1}{2}$ optimal. 

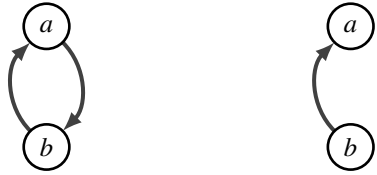
\begin{center}
\begin{tikzpicture}
    \Vertex[x=-2,y=0.8,label=$a$,fontsize=\normalsize,color=white]{a}
    \Vertex[x=-2,y=-0.8,label=$b$,fontsize=\normalsize,color=white]{b}
        \Vertex[x=2,y=0.8,label=$a$,fontsize=\normalsize,color=white]{a2}
    \Vertex[x=2,y=-0.8,label=$b$,fontsize=\normalsize,color=white]{b2}
\Edge[Direct,bend=45](a)(b)
\Edge[Direct,bend=45](b)(a)
\Edge[Direct,bend=45](b2)(a2)
\end{tikzpicture}
\end{center}

The above example relies on the case of two participants who are allowed to abstain. Several papers have circumvented the ceiling of $\frac{1}{2}$ by ruling out this case with conditions on the profile of nominations. If participants cannot abstain, the permutation mechanism achieves an approximation ratio of at least $\frac{7}{12}$ \citep{fischer2014optimal,fischer2015optimal}. If each participant submits exactly one nomination, the permutation mechanism has an approximation ratio of $\frac{2}{3}$ \citep{cembrano2023improved}. If the participant with the most nominations receives at least a threshold number of nominations, \citet{bousquet2014near} design a ``slicing mechanism" that has an approximation ratio close to one. The slicing mechanism first samples some participants to decide how the remaining participants should be partitioned and the order in which the partitions should be considered. The slicing mechanism adds a sampling step to the techniques used in partition and permutation mechanisms. The nearly optimal approximation ratio of the slicing mechanism relies on placing conditions on the nominations---the input to the mechanism. We can also consider how conditions on the output of the mechanism impact the approximation ratio.

Similar to how flexibility in the prizes provides mechanisms with better axiomatic properties, flexibility in the prizes also allows for better approximation ratios. If the mechanism targets $k$ winners, \citet{bjelde2017impartial} show that allowing for the mechanism to select fewer than $k$ winners in some situations can allow for improved approximation ratios. In comparison to the case where the mechanism must select exactly two winners, allowing the mechanism to sometimes select fewer winners can improve the approximation ratio from $\frac{7}{12}$ to $\frac{2}{3}$. 

The approximation results described above all focus on the approximation ratio. Why should the mechanism designer target the ratio and not some other metric? \citet{caragiannis2019additive} propose additive approximation. Instead of the ratio, the designer targets the difference between the winner and the participant with the most nominations. For example, suppose the winner receives 5 nominations, but the participant with the most nominations receives 8 nominations. The approximation ratio is $\frac{5}{8}$ whereas the difference is 3. In the case of single nominations and no abstentions, a randomized partition mechanism provides an additive approximation of $\mathcal{O}(\sqrt{n})$ \citep{caragiannis2019additive}. A deterministic threshold mechanism can also achieve an additive approximation of $\mathcal{O}(\sqrt{n})$ \citep{cembrano2022additive}.

The study of the additive approximation ratio by \citet{caragiannis2019additive} and \citet{cembrano2022additive} assumes the mechanism can select at most one winner. If the mechanism is allowed to select many winners, the additive approximation ratio must be defined differently. \citet{cembrano2022correspondences} propose that the number of nominations for the participant with the most nominations should be compared to the selected participant with the least nominations. For example, suppose the most popular participant received 8 nominations, and the mechanism selected two participants, one with 7 nominations and another with 5 nominations. The min-additive approximation proposed by \citet{cembrano2022correspondences} is $8 - 5 = 3$. The plurality with runners-up mechanism proposed by \citet{tamura2014impartial} is 1-min-additive, which turns out to be the best possible min-additive approximation \citet{cembrano2022correspondences}. 

The approximation results described above all focus on the participant with the most nominations. In social networks, the mechanism designer may wish to target the most influential user rather than the user who is the most popular. In social networks, participant $a$ nominating participant $b$ can be thought of as user $a$ following user $b$. If we consider the graph of nominations, the approximation results discussed above target the participant with the maximum in-degree, but other measures of centrality may more accurately capture influence. For example, \citet{babichenko2018diffusion} define an influence measure using the expected number of paths that will end at a given participant when starting randomly at any participant. 

Similar to the way participants can manipulate which participant receives the most nominations, participants can also manipulate which participant has the highest influence measure. Consider the example shown below. Participant $a$ and $c$ both have two nominations (or ``follows" in the language of social media), but $a$ is more influential than $c$ because he can influence $d$ and $e$ via his influence on $c$. Suppose the mechanism selects $a$. If $c$ chooses to abstain and remove his nomination, $c$ would become the most influential participant. Participant $c$ can change the outcome by changing his nomination.

\begin{center}
\begin{tikzpicture}
    \Vertex[x=2,y=1,label=$a$,fontsize=\normalsize,color=white]{a}
    \Vertex[x=1,,y=0,label=$b$,fontsize=\normalsize,color=white]{b}
    \Vertex[x=3,y=0,label=$c$,fontsize=\normalsize,color=white]{c}
    \Vertex[x=2,y=-1,label=$d$,fontsize=\normalsize,color=white]{d}
    \Vertex[x=4,y=-1,label=$e$,fontsize=\normalsize,color=white]{e}
\Edge[Direct](b)(a)
\Edge[Direct](c)(a)
\Edge[Direct](d)(c)
\Edge[Direct](e)(c)
\end{tikzpicture}
\end{center}

\citet{babichenko2018diffusion} design mechanisms that are impartial and approximate the ideal of selecting the most influential participant. When the nomination graph is a tree or a forest, an impartial mechanism with a constant approximation ratio to the maximum influence exists. \citet{babichenko2020forests} provide tighter bounds on the approximation for forests, \citet{zhang2021} improve the bounds for directed acyclic graphs when participants can only manipulate by hiding nominations, and \citet{zhao2023} design a mechanism that achieves the upper bound shown in \citet{zhang2021}.

\FloatBarrier

\begin{table}
\thisfloatpagestyle{empty} 
\caption{Empirical evidence of peer mechanisms}
\label{table:empirics}
\centerfloat
\small
 \begin{tabular}{m{33mm} m{16mm} m{34mm} m{40mm} m{28mm} m{10mm}}
 \toprule
 Paper	&	Setting	&	Context	&	Input	&	Participants	&	Sample Size	\\
 \midrule
\citet{alatas2019}	&	Field	&	Government aid programs	&	Influence beneficiary lists	&	Residents	&	3998	\\
\citet{alatas2016}	&	Field	&	Government cash transfers	&	Rank 8 households	&	Residents	&	5633	\\
\citet{hussam2021targeting}	&	Field	&	Business grants	&	Rank 5 entrepreneurs	&	Entrepreneurs	&	1345	\\
\citet{huang2019discovery}	&	Field	&	Employee promotion	&	Grade peers	&	Employees	&	432	\\
\citet{trachtman2021}	&	Field	&	Cash transfers	&	Rank 10 households	&	Residents	&	300	\\
\citet{dupas2022poverty}	&	Field	&	Poverty targeting	&	Rank neighbors	&	Residents 	&	507	\\
\citet{carpenter2010}	&	Lab	&	Worker compensation	&	Grade 7 peers	&	Students	&	224	\\
\citet{balietti2016peer}	&	Lab	&	Art exhibition	&	Grade 3 peers	&	Students	&	144	\\
\citet{chakraborty2018}	&	Lab	&	Peer grading	&	Grade 5 peers	&	Students	&	69	\\
\citet{leibbrandt2018gender} & Lab & Neutral & Grade 3 peers & Students & 200 \\
\citet{kotturi2020hirepeer}	&	Lab	&	Freelance job applications	&	50 pairwise comparisons	&	MTurk workers	&	320	\\
\citet{stelmakh2020catch}	&	Lab	&	Neutral	&	Rank 4 peers	&	Students	&	55	\\
\citet{bao2021deterrence}	&	Lab	&	Crime	&	Nominate 1 peer	&	Students	&	300	\\
\citet{piech2013mooc}	&	Observational	&	Peer grading	&	Grade 4 peers	&	Students	&	3600	\\
\citet{basurto2020}	&	Observational	&	Government subsidies	&	Nominate households	&	Residents	&	1559	\\
\citet{vera2022}	&	Observational	&	Government loans	&	Assess loan applications	&	Residents	&	710	\\
 \bottomrule
 \end{tabular}
\end{table}

\section{Empirical Evidence}\label{sec:empirics}

In this section, we discuss the empirical evaluation of peer mechanisms. After providing an overview of the studies and discussing evidence of manipulation, we highlight several key lessons the empirical studies provide for theory.

Table \ref{table:empirics} lists research studies that test peer mechanisms in practice. We focus on studies where the participants providing the peer reports are also eligible for the prize. Many fascinating studies that ask a third party to report on the participants are excluded.%
\footnote{For example, \citet{maitra2020} study mechanisms where local traders and political representatives nominate farmers to receive loans. We do not include \possessivecite{maitra2020} study in Table \ref{table:empirics} because the local traders and political representatives are not eligible to receive loans.}

We separate studies into three settings: field experiments, lab experiments, and observational studies. The field experiments introduce experimental treatments in real-life settings. For example, \citet{hussam2021targeting} assigned business grants to entrepreneurs in India based on peer rankings of profitability. The lab experiments invite participants into a controlled laboratory setting and study the impact of experimental treatments. Lab experiments may be conducted in physical locations or online. \citet{carpenter2010} recruit student participants to complete tasks in a laboratory environment while \citet{kotturi2020hirepeer} conduct experiments using the Amazon Mechanical Turk online crowdsourcing platform. The line between field and lab experiments can be unclear. As \citet{chakraborty2018} point out, their lab experiment studies peer grading in a classroom setting so could be considered a field experiment. The final setting is observational studies, where the researchers did not introduce any experimental treatments.

As peer mechanisms can have many different applications, the studies listed in Table \ref{table:empirics} have been conducted in many different contexts. Many studies focus on the context of government transfers, subsidies, and loans. Policymakers have recognized that local community members may have superior information compared to the central government on which community members are in most need of aid or will make the most productive use of a loan. Mechanisms that rely on the local community to target aid (often called ``community-based targeting") are a popular means to decide which community members should receive aid.

Another popular study context is peer grading of student assignments. As online delivery has enabled instructors to scale their courses to thousands of students, the instructor does not have the time to grade all the assignments. Peer grading offers a scalable solution for grading in massive open online courses (MOOCs). Although there are many studies on different facets of peer grading, we focus on studies that highlight how the manipulation of grades can be prevented.

Outside of the targeting of government programs and peer grading of student assignments, empirical studies of peer mechanisms have focused on a diverse range of contexts. \citet{huang2019discovery}, \citet{carpenter2010}, and \citet{kotturi2020hirepeer} focus on labor markets: peer evaluations can influence which employee is promoted, set salary bonuses, or decide which freelancer is hired. \citet{balietti2016peer} use peer grading to assess artworks. \citet{bao2021deterrence} use a lab experiment to show how allowing criminal suspects to nominate another suspect can reduce the overall crime level. \citet{hussam2021targeting} use peer evaluations to determine which entrepreneurs will create the largest return from a business grant or loan.

We also note that Spliddit, a popular online tool for using fair division algorithms, has implemented \possessivecite{declippel2008impartial} peer mechanism to divide credit for a joint project among the members of the team \citep{goldman2015spliddit}. Empirical studies of Spliddit have focused on other functionality, such as the rent sharing algorithms \citep{gal2017spliddit}, and the peer mechanism part of the tool has not been the main focus. Spliddit's peer mechanism is a small part of \possessivecite{lee2017spliddit} user study that focuses primarily on the algorithms for assigning chores. The study did not provide insights on preventing the manipulation of peer mechanisms.

The empirical studies listed in Table \ref{table:empirics} use the full range of inputs discussed in the taxonomy in Section \ref{sec:taxonomy}: nominations, rankings, and grades. Except for \citet{kotturi2020hirepeer}, most studies asked participants to evaluate a small set of peers for nominations, rankings, and grades. Evaluating a large number of peers is likely to be tedious or cognitively demanding. However it is unclear how many peers people can evaluate before accuracy begins to erode. Participants may also find it easier to use nominations than ranking or grading as the number of peers increases.

In the final two columns of Table \ref{table:empirics}, we list the type of study participants and sample size. The sample sizes range from large countrywide studies to small lab experiments.

\subsection{Evidence of manipulation}

Although peer mechanisms can create opportunities to manipulate who wins the prize, do participants take these opportunities? The following examples show that they do.
\begin{itemize}
\item In the context of employee promotion, \citet{huang2019discovery} shows that employees will reduce peer grades for coworkers eligible for the same promotion. If their coworker was not eligible and therefore not a direct competitor, employees tended to inflate their peer grade of the coworker.
\item In a field experiment with entrepreneurs in India, \citet{hussam2021targeting} show that peer reports decrease substantially in accuracy when the reports influence the chance of receiving a business grant.
\item During a lab experiment conducted by \citet{carpenter2010}, participants were asked to print, seal, and address letters to a list of recipients---complete with handwritten addresses. Each participant was then asked to count the number of letters and rate the quality of the work of each of their seven peers in the experiment. When the peer reports determined who received a bonus, the participants under-counted the number of letters and reduced quality ratings. Participants relied more on the subtle manipulation of reducing the quality rating than the more obvious manipulation of under-counting.
\item During a lab experiment framed as an art competition, participants gave lower peer review scores to direct competitors than to other peers when the prize was split among the winners \citep{balietti2016peer}. When all winners received the same prize and the number of winners was unlimited, participants gave similar scores to direct competitors and other peers. 
\end{itemize}

As these examples show, participants in peer mechanisms do tend to take opportunities to manipulate peer mechanisms in their favor. The examples highlight one type of manipulation---downgrading competitors---but manipulation can take other forms, such as collusion and nepotism.

Notorious cases in academic peer review provide examples of collusion \citep{ferguson2014scam,littman2021collusion}. One researcher gives another researcher a positive review on their paper in exchange for a positive review in return. The collusion can be more complex than two researchers exchanging positive reviews. A collusion ring may form where researcher a larger group of researchers exchange reviews.\footnote{\citet{jecmen2024detection} provide evidence that collusion rings are difficult to detect.} For example, in a collusion ring with three researchers, $a$ writes a positive review about $b$, then $b$ writes a positive review about $c$, and the ring closes with $c$ writing a positive review about $a$. 

For an example of nepotism, entrepreneurs in \possessivecite{hussam2021targeting} study tended to increase the rank of friends and family members---especially when the peer rankings influenced the chance of receiving a business grant.

\subsection{Lessons for theory}\label{sec:lessons_for_theory}

The empirical research provides several key lessons for the theoretical analysis of peer mechanisms. We recognize that theoretical models should not be too complex, so we encourage researchers to choose carefully which of these lessons, if any, to include in their models.

\subsubsection{Participants make errors in their peer evaluations}

Across different contexts, participants make errors in their peer evaluations. In peer grading of student assignments, some students consistently inflate or deflate the grades they give peers, and students differ in the reliability of their grades \citep{piech2013mooc}. In the context of poverty targeting, community members often report that they don't know the ranking of two fellow community members \citep{alatas2016}. 

If the mechanism designer can assume that participants do not make errors, she can design a mechanism that punishes differences in peer evaluations. If two participants share different evaluations of a given peer, one must be lying. The chance of errors makes such mechanisms difficult to implement. The designer does not know whether the participant is lying or making an honest error. Any mechanism that punishes differences or rewards consensus must consider the chance of errors.

In some contexts, participants may hold very little accurate information about their peers. For example, \citet{dupas2022poverty} asked survey respondents in Abidjan, C\^{o}te D'Ivoire to rank neighbors from poorest to richest. The ranking contained many cycles and had a weak correlation with other measures of wealth. An open problem is to design a mechanism that can adjust to the level of peer information participants hold. Perhaps the mechanism could award more prizes or larger prizes when participants provide more accurate peer evaluations. 

A lack of consensus in peer evaluations is a clear sign that the participants have made errors. But, the participants may make errors even when they agree. \citet{trachtman2021} found that although residents agreed on peer rankings of need, the rankings reflected long-term attributes, and the residents could not identify which of their fellow residents were in immediate need. If the mechanism designer aimed to collect a ranking of immediate need, the consensus peer rankings would not reflect this aim.

\subsubsection{Nepotism is common}

The empirical research shows several examples where participants favor family members and friends in peer evaluations. Entrepreneurs gave higher ranks to family members and friends when asked to rank fellow entrepreneurs according to profitability \citep{hussam2021targeting}. Village chiefs were more likely to give food subsidies to family members \citep{basurto2020}. A committee assigning local business loans was more likely to give loans to community members who were socially connected to the committee \citep{vera2022}.

Most models of peer mechanisms consider that participants treat all peers equally. The empirical research shows that we cannot ignore the social context of peer mechanisms. Participants often favor their friends and family. Just as researchers have developed many creative approaches to discourage selfish manipulation, we also need creative approaches to discourage or detect nepotism.

\subsubsection{Small amounts of manipulation can be acceptable}

The same studies that have shown evidence of nepotism also emphasize that peer mechanisms can still be the best option even if they are susceptible to manipulation \citep{alatas2019,basurto2020}. A common alternative to a peer mechanism decides on winners based on participants' applications sent to an external committee. This more centralized approach may be more costly and less effective than a peer mechanism---even when participants have manipulated the outcome of the peer mechanism.

Most models of peer mechanisms try to find equilibria where all participants report the truth.\footnote{\citet{baumann2018self} is one exception. The mechanism uses an equilibrium where participants misreport their peer evaluations.} Allowing for small amounts of manipulation may allow researchers to take new approaches to designing peer mechanisms.

\subsubsection{Choosing optimal manipulation can be difficult}

During a lab experiment conducted by \citet{stelmakh2020catch}, participants were encouraged to manipulate their peer rankings. The experiment showed that many participants struggled to employ the optimal manipulation. Participants often chose to reverse their reported peer ranking, which did not help to move themselves up the final ranking. Reversing the ranking could make the participant worse off if the participant was near the bottom of the ranking by boosting the score of their closest competitors (peers also near the bottom of the ranking). 

The optimal manipulation put peers closest to the participant at the bottom of the ranking and peers furthest away at the top. With some nudging, most participants employed the optimal manipulation after a few rounds of practice playing against truthful bots. 

A layer of complexity was added when playing against participants who could also manipulate their peer reports. Each participant needed to reason about the types of manipulations their peers might employ. The optimal manipulation depends on their peers' actions.

Peer mechanisms could take advantage of the complexity of finding optimal manipulations. Participants may prefer to report truthfully if they are uncertain about the optimal manipulation or cannot calculate the optimal manipulation. \citet{conitzer2016handbook} survey how computational complexity can be a barrier to manipulation in voting---similar insights may apply to peer mechanisms.

\section{Research Challenges}

Based on our reading of the theoretical and empirical research on peer mechanisms, we highlight several important research challenges.

\subsection{Collusion}\label{ch:collusion}

If peers can communicate, they can collude. ``I will give you a positive review if you give me a positive review.'' Peer mechanisms must prevent manipulation by groups as well as by individuals.

Most existing peer mechanisms can be manipulated by groups. One partial exception is the partition approach. Reviewers placed within the same group cannot collude because their reviews only impact peers outside of their group. Unfortunately, the partition approach cannot prevent collusion between participants in different groups.

Preventing all forms of collusion is likely impossible. For example, in the model of nominating one or more peers for a fixed number of winners, \citet{alon2011sum} prove the impossibility of designing a group-strategy-proof peer mechanism with good performance. Even if we cannot design peer mechanisms immune to collusion, the challenge of discouraging and detecting collusion is still important.

At a minimum, the mechanism should not encourage collusion.%
\footnote{For example, in designing algorithms to assign wine producers to certify fellow producers, \citet{barrot2020wine} recognize that if two producers are assigned to review each other, they may be tempted to collude. The algorithms include a constraint that two producers cannot review each other.} 
Consider a simple peer mechanism for poverty targeting---give aid to a person if he claims he is poor and his neighbor agrees. The mechanism is impartial but provides a strong incentive to collude. Suppose the claimant is rich. He could claim he is poor and pay his neighbor to agree by giving his neighbor a portion of the aid.

As \citet{rai2002targeting} shows, the pressure to collude can be alleviated by limiting the aid budget. If two neighbors both claim to be poor, they each receive half the budget, whereas if they claim that one of the neighbors is rich, the poor neighbor receives the full budget. The disadvantage is that poor claimants with rich neighbors receive less aid than poor claimants with poor neighbors.

Besides early work by \citet{rai2002targeting} on poverty targeting, we know little about how to discourage collusion in peer mechanisms, in which settings collusion is most likely, or how to detect collusion. We encourage work on these open questions.


\subsection{Nepotism}\label{ch:other_preferences}

Much research on peer mechanisms starts with the assumption that people only care about their own chance of winning the prize. If it's not me, then I don't care who wins.

Academic peer review provides a counterexample to the assumption. Reviewers are often asked to list conflicts of interest—to avoid biases towards colleagues, coauthors, and students. The conflicts of interest are usually public knowledge and can be modeled as a conflict graph. The mechanism designer can prevent manipulation by choosing a partition that respects the conflict graph \citep{xu2019conflictgraph}.  

What if the conflicts of interest are not public knowledge? As discussed above, empirical research shows many examples of nepotism. Participants often favor family and friends. The mechanism designer may not observe these social connections. Can the mechanism discourage nepotism without observing the conflicts of interest? \citet{hussam2021targeting} show that using peer prediction mechanisms to pay for accuracy can discourage nepotism. Are there other approaches that can discourage nepotism?

\subsection{Punishments not prizes}

Most contexts that motivate peer mechanisms, such as peer grading and awarding research grants, involve awarding a prize. The participants want to be selected. However, peer mechanisms can also be used for assigning chores or punishments \citep{bao2021deterrence}. In this case, the participants want to avoid being selected.

Is the problem of assigning prizes equivalent to the problem of assigning punishments? Or are there differences that allow for different types of mechanisms? In fair division, the allocation of chores is somewhat different to the allocation of goods (e.g. \cite{aciwijcai2019}). We expect peer mechanisms for punishments to behave somewhat differently to peer mechanisms for prizes.


\subsection{Machine learning methods}\label{ch:ML}

We might view this not as the problem of designing a peer mechanism with good axiomatic properties but as a machine learning task, with a focus that shifts to accuracy and error. We might also view the problem of identifying manipulation (and colluding peers) as a prediction problem suitable for machine learning.  For example, reinforcement learning has been proposed to prevent collusion in online e-commerce platforms \citep{brero2022learning}. And similar to how deep learning has been used to design optimal auctions \citep{dutting2019optimal}, we might be able to employ machine learning methods to learn peer mechanisms with good properties. 


\subsection{Mechanisms that can respect constraints}\label{ch:constraints}

We might have constraints on the winners. For example, we might want a gender-balanced group of winners.  One solution to this problem is to have the women vote on the male winners and the men on the female winners. However, this may not be very satisfactory if the women know more about each other and the men similarly. How then can we adapt the peer mechanisms discussed so far to deal with additional constraints like this? Such constraints have proved useful for capturing real-world issues in other areas of social choice (e.g. diversity constraints in school choice mechanisms \citep{aziz2019matching}), and we may be able to adapt ideas from these domains to peer mechanisms. 

Peer mechanisms that respect constraints can have unintended consequences that create new challenges. A lab experiment studied how a gender quota impacted groups of two men and two women completing tasks for payment \citep{leibbrandt2018gender}. In one version, performance was measured by peer review, and only the top two performing participants in each group received higher pay. Participants could manipulate the mechanism by under-reporting the number of tasks their peers completed. When a gender quota was in place (where at least one woman received higher pay), peers were more likely to under-report the performance of women. The difference was driven by women being more likely to sabotage women in the presence of a gender quota while men sabotaged women and men equally. The gender quota had the unanticipated consequence of intensifying competition between women and focusing women's manipulation of peer review on other women.

\section{Conclusion}

Manipulation is a very real problem in peer mechanisms where a  group is selecting one or more of the group to win a prize,  receive a ranking, or be given a grade. This survey identified three broad approaches to prevent such manipulation: mechanisms designed to be impartial so that a participant cannot impact their outcome, audits to detect and punish manipulation, and rewards for truthful reports. Empirical evidence of manipulation in practice suggests several outstanding research challenges, such as dealing with collusion between participants as well as various forms of nepotism. Despite the considerable body of research in this area, there remain many significant obstacles to be overcome in the design of peer mechanisms to address a range of issues met in the real world. 

\section*{Acknowledgments}

This research was supported under Australian Research Council's Laureate Fellowship (project number FL200100204). We thank Felix Fischer, Andrew Mackenzie, Herve Moulin, Axel Niemeyer, Shinji Ohseto, and Nihar Shah for their helpful suggestions. We also thank the anonymous referees for the detailed comments that helped to improve the paper.    

\bibliographystyle{elsarticle-harv}
\bibliography{references}


\end{document}